\documentclass{article}

\PassOptionsToPackage{numbers,sort&compress,comma,square}{natbib}

\usepackage{ifthen}
\newboolean{submission}
\setboolean{submission}{false} 

\setboolean{submission}{true}
\usepackage[preprint]{neurips_2024} \setboolean{submission}{false}

\usepackage{graphicx}
\usepackage{booktabs} 
\usepackage{tikz}
\usepackage{pgfplots}
\pgfplotsset{compat=1.17}
\usepackage{float} 
\usepackage{amsmath}
\usepackage[utf8]{inputenc} 
\usepackage[T1]{fontenc}    
\usepackage{hyperref}       
\usepackage{url}            
\usepackage{amsfonts}       
\usepackage{nicefrac}       
\usepackage{microtype}      
\usepackage{xcolor}         
\usepackage{subcaption}
\usepackage[utf8]{inputenc}
\usepackage{tcolorbox}
\usepackage{enumitem}
\usepackage{amsmath}
\usepackage{booktabs}
\usepackage{wrapfig}

\definecolor{set2-1}{RGB}{102,194,165}
\definecolor{set2-2}{RGB}{252,141,98}
\definecolor{set2-3}{RGB}{141,160,203}
\definecolor{set2-4}{RGB}{231,138,195}
\definecolor{set2-5}{RGB}{166,216,84}
\definecolor{set2-6}{RGB}{255,217,47}
\definecolor{set2-7}{RGB}{229,196,148}
\definecolor{set2-8}{RGB}{179,179,179}

\usepackage{tabularx}

\title{Revisiting Multi-Modal LLM Evaluation}

\author{Jian Lu$^{1,}$\thanks{Corresponding author: jlu59@u.rochester.edu} \qquad Shikhar Srivastava$^{1}$ \qquad Junyu Chen$^{1}$ \qquad \textbf{Robik Shrestha}$^{1}$ \\ \textbf{Manoj Acharya}$^2$ \qquad \textbf{Kushal Kafle}$^3$ \qquad \textbf{Christopher Kanan}$^{1}$\\
$^1$University of Rochester \qquad $^2$SRI International \qquad $^3$Adobe\\
}

\begin{document}

\maketitle

\begin{abstract}
With the advent of multi-modal large language models (MLLMs), datasets used for visual question answering (VQA) and referring expression comprehension have seen a resurgence. However, the most popular datasets used to evaluate MLLMs are some of the earliest ones created, and they have many known problems, including extreme bias, spurious correlations, and an inability to permit fine-grained analysis. In this paper, we pioneer evaluating recent MLLMs (LLaVA 1.5, LLaVA-NeXT, BLIP2, InstructBLIP, GPT-4V, and GPT-4o) on datasets designed to address weaknesses in earlier ones. We assess three VQA datasets: 1) TDIUC, which permits fine-grained analysis on 12 question types; 2) TallyQA, which has simple and complex counting questions; and 3) DVQA, which requires optical character recognition for chart understanding. We also study VQDv1, a dataset that requires identifying all image regions that satisfy a given query. Our experiments reveal the weaknesses of many MLLMs that have not previously been reported. Our code is integrated into the widely used LAVIS framework for MLLM evaluation, enabling the rapid assessment of future MLLMs. Project webpage: \url{https://kevinlujian.github.io/MLLM_Evaluations/}
\end{abstract}

\section{Introduction}
In recent years, multi-modal large language models (MLLMs) have emerged as powerful tools for tackling vision-language tasks~\cite{li2023blip,chowdhery2022palm,zhu2023minigpt,koh2023grounding,liu2023llava}. Open source MLLMs leverage the extensive world knowledge of large language models (LLMs) and combine them with pre-trained vision encoders to process both linguistic and visual information~\cite{liu2023llava,zhu2023minigpt,liu2023improved}. These models are trained on various vision-language tasks such as visual question answering (VQA)~\cite{goyal2017making,zhang2016yin}, image captioning~\cite{sharma2018conceptual}, and visual conversations~\cite{sharegpt}. Their effectiveness is typically evaluated on VQA datasets~\cite{goyal2017making,cocoqa}, which test the ability to produce answers to questions about images and referring expression comprehension tasks~\cite{refcocokazemzadeh2014referitgame}, which require localizing the single object specified in the referring expression.

From 2017-2019, our lab created a series of datasets intended to enable fine-grained analysis of visually grounded language understanding systems: 
\begin{enumerate}[noitemsep,nolistsep]
    \item \textbf{VQDv1}~\cite{Acharya2019VQD}, which requires the model to produce multiple bounding boxes instead of localizing only one object, thereby testing for general query detection skills;
    \item \textbf{TallyQA}~\cite{Acharya2018TallyQA}, which tests visual grounding through counting skills, asking questions that require intricate reasoning;
    \item \textbf{TDIUC}~\cite{kafle2017analysis}, which tests versatility across 12 tasks, including object, attribute, and activity recognition, as well as overall scene understanding; and 
    \item \textbf{DVQA}~\cite{kafle2018dvqa}, which requires interpreting and analyzing visual data in chart form, testing for the ability to do OCR, and properly handling unusual words found in charts.
\end{enumerate}
While our datasets were widely used for evaluating VQA systems prior to the MLLM era, surprisingly, they have not been used to evaluate MLLMs. Based on community feedback, this is for two reasons: 1) The test datasets are prohibitively large, despite enormous improvements in GPUs, and 2) They are not integrated into widely used toolboxes for evaluating MLLMs, such as LAVIS. Here, we address these issues by creating ``slim'' versions of the datasets, and we integrate them into LAVIS. We then benchmark recent MLLMs on them, including GPT-4o~\cite{gpt4o}.

Our datasets are designed to overcome the widely acknowledged weaknesses of earlier datasets\cite{cocoqa,goyal2017making,refcocogmao2016}. Despite this, these early datasets are now widely used to evaluate MLLMs. The most commonly used datasets, e.g. VQAv2\cite{goyal2017making}, fail to adequately gauge visual grounding, allowing models to inflate performance by exploiting language bias without using visual information~\cite{Kafle2016AnswerType}. Additionally, they do not categorize questions into types, preventing fine-grained analysis of abilities like attribute detection, object recognition, reasoning, and scene understanding. In contrast, TDIUC provides comprehensive evaluation across 12 diverse tasks, enabling fine-grained analysis, while TallyQA focuses on counting, demanding intricate spatial reasoning for its complex questions. DVQA challenges models with chart interpretation, requiring OCR and handling unusual words. Referring expression datasets like RefCOCO\cite{refcocogmao2016} often only require localizing a single object, allowing models to exploit biases~\cite{cirik2018visual,Acharya2019VQD} and often can answer queries without even considering the sentence structures \cite{akula2020words}. In contrast, VQDv1 requires identifying multiple objects or none based on the query, making it a more rigorous test for visual grounding and reducing the ability to exploit biases.

\paragraph{This paper makes the following contributions:}
\begin{enumerate}[noitemsep,nolistsep]
    \item We create new ``slim'' versions of the datasets suitable for zero-shot MLLM evaluation, and they are integrated into the widely-used LAVIS toolbox, facilitating the rapid and comprehensive assessment of future MLLMs.
    \item We provide a robust evaluation of MLLMs on our VQA datasets, revealing previously unreported weaknesses via fine-grained analysis across various question types and tasks.
    \item Using VQDv1, we challenge MLLMs' visual grounding capabilities by requiring them to engage in complex visual reasoning to identify multiple objects beyond the limitations of single-object referring expression datasets.
\end{enumerate}

\section{Multi-modal Large Language Models}

Open-source MLLMs comprise a pre-trained LLM, a pre-trained vision encoder, and a learned adapter that aligns the visual and linguistic representations~\cite{zhu2023minigpt,liu2024visual}. They are usually trained in multiple stages. Initially, the adapter is trained to align the visual embeddings generated by the vision encoder with the textual embedding space of the LLM. Subsequently, the MLLM undergoes fine-tuning by adapting both the adapter and the LLM on various vision-language and instruction-tuning datasets. In our study, we consider both widely available state-of-the-art open-weight MLLMs and closed-source MLLMs.

\textbf{BLIP2}~\cite{li2023blip} is a generic and compute-efficient method for vision-language pre-training that leverages frozen pre-trained image encoders and language models (LLMs). It pre-trains a lightweight Querying Transformer (Q-Former), consisting of image and text transformer sub-modules, to bridge visual and textual modalities. BLIP2, therefore, only trains a relatively light - 188M parameter transformer and achieves strong performance on VQA and image captioning tasks. We evaluate the base BLIP2 model~\cite{li2023blip}, with {`blip2-flan-t5-xl'} as the pretrained encoder.

\textbf{iBLIP}~\cite{dai2024instructblip} (i.e., InstructBLIP), like BLIP-2, keeps the LLM and visual encoders frozen while introducing a novel instruction-aware Query Transformer that allows the model to extract informative visual features based on the textual instructions in the prompt. iBLIP is additionally trained on a much larger corpus of visual instruction tuning datasets, including knowledge-grounded image-question answering, visual reasoning, and VQA~\cite{dai2024instructblip}. This leads to improvements, including higher zero-shot performance on VQA tasks, compared to BLIP2 and larger MLLMs. We test the version that uses `instructblip-flan-t5-xxl' as the pre-trained encoder.

\textbf{LLaVA}~\cite{liu2023llava} uses a visual instruction tuning dataset to fine-tune the LLM and adapter. LLaVA 1.5 enhances its vision encoder to handle higher-resolution images and replaces the linear projector layer with a multi-layer perceptron adapter. This version is trained on the VQA datasets VQAv2 and GQA datasets and a broader range of instruction-tuning data from sources like ShareGPT. These enhancements significantly improve its performance on fine-grained visual tasks, including detailed image description and complex question answering ~\cite{liu2023improved}. It achieves strong performance on several VQA benchmarks. 

\textbf{LLaVA-NeXT}~\cite{liu2024llavanext} is an improved version of LLaVA 1.5, with a focus on enhanced visual reasoning, optical character recognition (OCR), and multi-modal document understanding. LLaVA-NeXT scales the input image resolution of input images by 4$\times$, up to \(1344 \times 336\) compared to \(336 \times 336\) in LLaVA 1.5 to enhance its ability to grasp finer-grained visual cues. LLaVA-NeXT is also trained on a more diverse and realistic visual instruction-tuning dataset (ShareGPT-4V and LAION-GPT-V), as well as a range of OCR, document, and chart datasets. We evaluate the 7B parameter version of LLaVA-NeXT.

\textbf{GPT-4o/GPT-4V~\cite{achiam2023gpt,yang2023dawn}} are closed-weight MLLMs created by OpenAI that enable users to leverage the capability of GPT-4 scale LLMs to analyze visual inputs. GPT-4V is a powerful generalist multi-modal model and can process arbitrarily interleaved image-text data. GPT-4V can perform many visual-language tasks well, including spatial understanding, object localization, and object counting~\cite{yang2023dawn}. GPT-4o is reportedly an end-to-end text, vision, and audio multi-modal model, where multi-modal tokens are processed within the same network. GPT-4o has also been reported to improve linguistic and multi-modal understanding. Given that these are closed-source MLLMs, we use the API provided by OpenAI for our evaluations.

\section{Creating ``Slim'' Evaluation Sets}

We evaluate MLLMs on the entire validation set of TallyQA, which contains 38,589 questions. However, the other datasets are much larger, which makes it challenging to quickly and inexpensively evaluate MLLMs on them. To address this, we sample subsets from these datasets for evaluation. A uniform random sampling is suboptimal as these datasets have long-tailed distributions and sampling uniformly would result in discarding examples from the tail. Therefore, we adopt a stratified sampling approach for DVQA and TDIUC, where we also maintain as much answer variety as possible. Specifically, we first categorize the questions into fine-grained groups, defined by both the pre-defined types in the datasets (e.g., question types or difficulty levels) and their corresponding answers. We define $r$ as the sampling ratio and $k$ as the minimum number of samples from each group. For any large group, we uniformly sample an $r$ proportion of the entries. For smaller groups, if the size  $m$ is such that $m\cdot r$ is less than $k$, we sample $k$ entries. For groups even smaller than $k$, we use the entire group. The number of samples $m^\prime$ to be taken from group $|g_i|=m$ can be represented as follows:
$$
m_i^\prime =
\begin{cases}
m_i & \texttt{if } m_i \leq k \\
k & \texttt{if } m_i \cdot r < k \land m_i > k \\
\lceil m_i \cdot r \rceil & \texttt{if } m_i \cdot r \geq k

\end{cases}
$$

VQDv1 has a long-tail distribution regarding the number of bounding boxes per query, where queries with 0 or 1 box comprise almost 90\%  of the dataset. Our goal is to evaluate the MLLM's ability to generate a variable number of bounding boxes -- extending the evaluation scope beyond traditional referring expression comprehension datasets such as RefCOCO \cite{refcocogmao2016}, where all referring expressions are associated with only one bounding box. Therefore, we retained all the questions with more than one bounding box and randomly sampled queries corresponding to 0 or 1 bounding box. As seen in Table~\ref{tab:bounding_boxes}, this method effectively increases the ratio of questions with multiple bounding boxes.

Our sampling method preserves the most challenges samples present in the original dataset, ensuring a comprehensive evaluation while significantly reducing computational overhead. Summary statistics for the datasets are given in Table~\ref{table:dataset_sample}.

\begin{table}[t]
\centering
\caption{Summary statistics for the VQA and VQD datasets we study.}
\begin{tabularx}{\textwidth}{l>{\centering}Xrrr}
\toprule
\textbf{Dataset Name} & \textbf{\# of Categories} & \textbf{\# of Unique Answers} & \textbf{Original Size} & \textbf{Sampled Size} \\
\midrule
TDIUC~\cite{kafle2017analysis} & 12 & 562 &  538,868&  27,336\\
TallyQA~\cite{Acharya2018TallyQA} & 2 & 16 & 38,589 & 38,589 \\
DVQA~\cite{kafle2018dvqa} & 3 & 2113 & 580,557 &  29,025\\
VQDv1~\cite{Acharya2019VQD} & 5 & 24 & 190,174 &  37,057\\
\bottomrule
\end{tabularx}
\label{table:dataset_sample}
\end{table}

\section{Experiments}

Across datasets we compute both micro performance, i.e., where every example is weighted equally, and macro performance, where we average across the mean score for different question/query types.

\begin{figure}[t]
\centering
\begin{subfigure}[b]{0.38\textwidth}
    \centering
    \includegraphics[width=\textwidth]{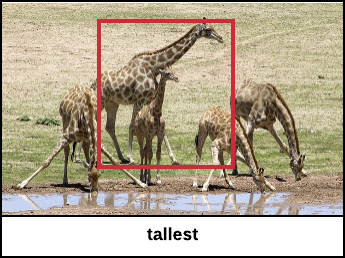}
    \caption{Referring Expression Comprehension}
    \label{fig:ref-exp-example}
\end{subfigure}
\hspace*{30pt}
\begin{subfigure}[b]{0.38\textwidth}
    \centering
    \includegraphics[width=\textwidth]{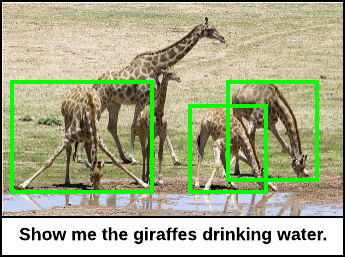}
    \caption{Visual Query Detection (VQD)}
    \label{fig:vqd-example}
\end{subfigure}
\caption{While only one object needs to be detected in popular referring expression comprehension datasets, VQDv1 requires identifying all regions that satisfy a query.}
\label{fig:vqd-dataset}
\end{figure}


\subsection{Visual Query Detection with VQDv1}

Visual query detection (VQD) requires a model to provide bounding boxes for $0$-$N$ visual objects in response to a given query~\cite{Acharya2019VQD}. It is significantly more challenging than referring expression comprehension, which requires only localizing a single object in a scene. VQD aligns more closely with typical human referring behavior, where it is common to refer to multiple objects simultaneously. Unlike VQA, VQD requires the model to ground responses in visual inputs, providing direct evidence of task completion.

We evaluated all models on VQDv1 except for BLIP2 and iBLIP, which failed to produce bounding boxes under the zero-shot setting. All models were prompted to answer with a list of bounding boxes. In the case of VQDv1, no single prompt worked universally well across all models. A fair comparison, therefore, required that we carefully select prompts for each model to achieve the best possible performance. To maximize fairness, we created a small set of prompts, and the most performative prompts for each model were selected. We discuss our prompt selection choices in Appendix~\ref{sec:prompt-selection}.

\paragraph{VQDv1 Metrics.} In \cite{Acharya2019VQD}, average precision using an intersection over union (IoU) of 0.5 was used for evaluation; however, that requires scores for each box, which are unavailable for MLLMs. Therefore, we compute each model's micro and macro mean $F_1$ scores, recall, and precision. The predicted box with the highest IoU above 0.5 is considered a true positive for each ground-truth box, whereas any remaining predicted boxes are false positives. If a query has no ground truth bounding boxes, then the $F_1$ score is set to 1 when the model outputs no boxes. Otherwise, it is set to 0. Due to the limited number of questions with four or more bounding boxes, we grouped them.

\paragraph{Results for VQDv1.}

\begin{table}[t]
\centering
\caption{Performance comparison of various multi-modal large language models on VQDv1 dataset.}
\label{tab:vqd-results}
\begin{tabular}{lccccc} 
\toprule
Metrics/Model & LLaVA~(7B) & LLaVA~(13B) & GPT-4V & GPT-4o & LLaVA-NeXT\\
\midrule
Micro $F_1$  & 25.06 & 20.96 & 21.17 & 25.33 & \textbf{27.01}\\ 
Macro $F_1$ & 19.87 & 16.81 & 16.54 & 21.01 & \textbf{21.84}\\
\bottomrule
\end{tabular}
\end{table}

\begin{figure}[t]
    \centering

    \begin{subfigure}[b]{0.49\textwidth}
    \centering

    \includegraphics[width=\textwidth]{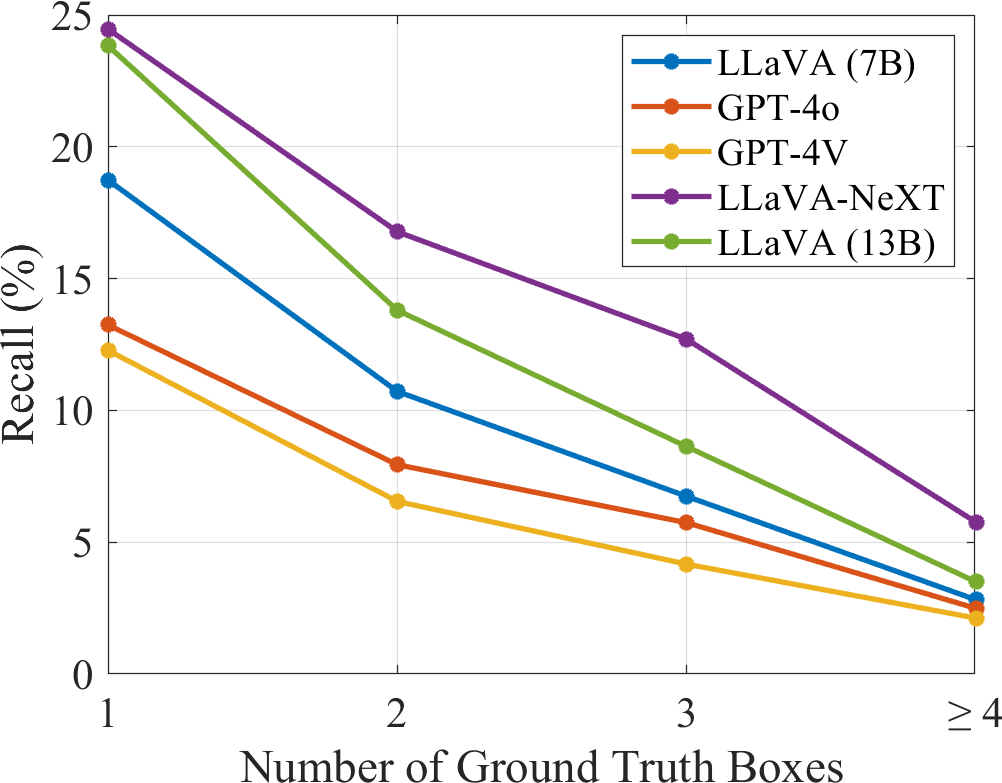}
    \caption{VQDv1 Recall}
    \label{fig:vqd-recall}
\end{subfigure}
\hfill
\begin{subfigure}[b]{0.49\textwidth}
    \centering
    \includegraphics[width=\textwidth]{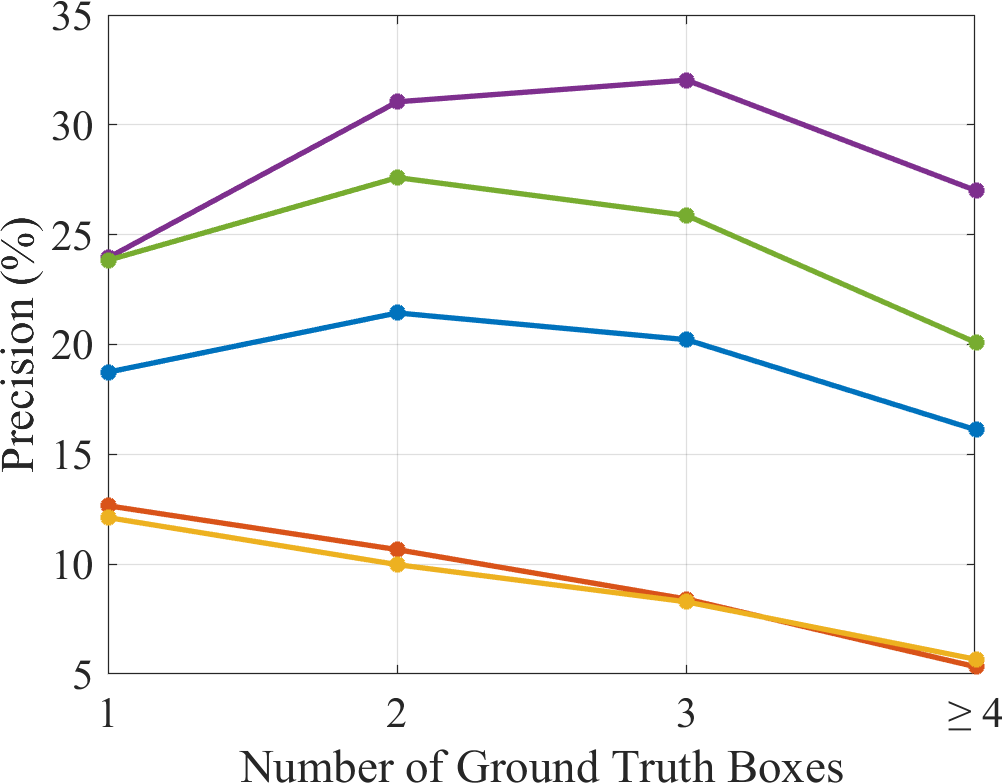}
    \caption{VQDv1 Precision}
    \label{fig:vqd-precision}
\end{subfigure}

    \caption{Recall and precision curves for queries with varying box counts.}
    \label{fig:vqd-per-box}
\end{figure}
As presented in Table~\ref{tab:vqd-results}, all of the models struggle on VQDv1, with the best performing LLaVA-NeXT obtaining only 27.01 in terms of micro $F_1$ score. Fig.~\ref{fig:vqd-per-box} shows the recall and precision scores across varying numbers of bounding boxes. Models struggle to ground multiple boxes, as evidenced by the recall score which decreases with an increase in the number of boxes.

\subsection{Fine-Grained VQA Assessment with TDIUC}

\begin{figure}[t]
    \centering
    \includegraphics[width=\linewidth]{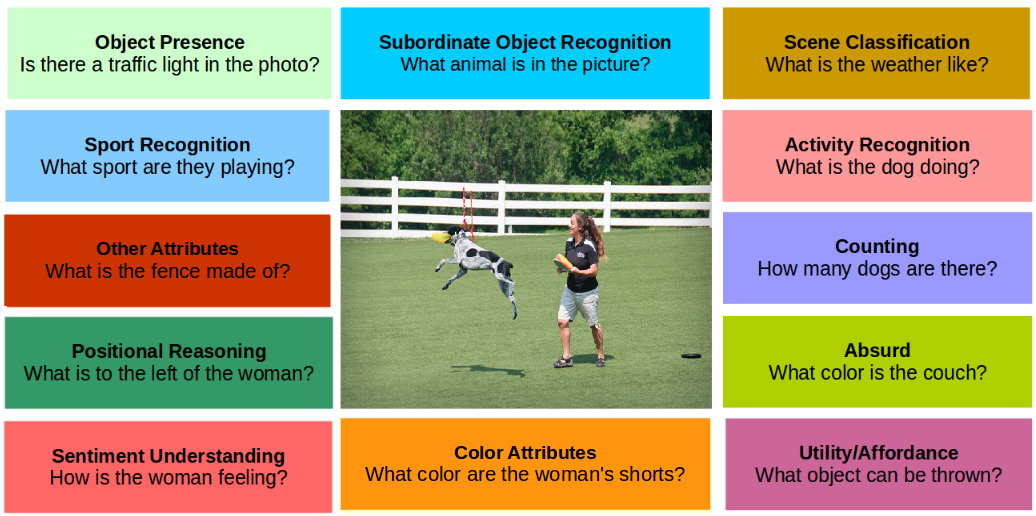}
    \caption{TDIUC has 12 kinds of questions, enabling fine-grained analysis of MLLMs.}
    \label{fig:tdiuc}
\end{figure}

TDIUC~\cite{kafle2017analysis} is a VQA dataset that organizes its questions into 12 distinct types as shown in Fig.~\ref{fig:tdiuc}. Performance is computed for each question type. TDIUC aims to address the shortcomings of previous VQA datasets by offering a broader spectrum of question types, and it enables a comprehensive analysis of VQA capabilities for each model. 

\paragraph{TDIUC Metrics.} For TDIUC, we use micro-accuracy and macro-accuracy, where micro accuracy corresponds to the average accuracy across the 12 question types. Macro-accuracy corresponds to the mean per type metric in the original paper.

\paragraph{Results for TDIUC.}
\begin{table}[t]
\centering
\caption{Accuracy on TDIUC for each question type. `L' denotes LLaVA. Best performers based on paired asymptotic McNemar tests ($\alpha = 0.05$) are in bold, except for Macro. Acc., where the max is bolded. For comparison, MuRel~\cite{Cadene2019} is the previous best result from training on TDIUC.}
\label{tab:model_performance_tdiuc}
\resizebox{\linewidth}{!}{
\begin{tabular}{lrrrrrrr|r}
\toprule
\textbf{Question Type} & \textbf{BLIP2} & \textbf{iBLIP} & \textbf{L (7B)} & \textbf{L (13B)} & \textbf{GPT-4V} & \textbf{GPT-4o} & \textbf{L-NeXT} & \textbf{MuRel}\\
\midrule
Absurd                  & \textbf{99.87} & 97.44 & 51.48 & 74.73 & 99.04 & 99.45 & 68.14 & 99.80\\
Activity Rec.         & 25.00 & 54.00 & \textbf{63.50} & 62.00 & 56.50 & \textbf{62.50} & \textbf{68.00} & 63.83\\
Attribute Rec.        & 1.31 & 48.15 & 71.46 & 73.20 & 60.78 & 73.20 & \textbf{79.08} & 58.19\\
Color Rec.            & 5.70 & 62.13 & 77.37 & \textbf{80.54} & 69.05 & 78.97 & \textbf{81.05} & 74.43\\
Counting                & 7.15 & 39.24 & 51.95 & 53.27 & 52.36 & \textbf{56.14} & \textbf{54.93} & 61.78\\
Object Pres.            & 43.22 & 74.87 & 91.31 & 90.57 & 67.28 & 77.81 & \textbf{92.07} & 95.75\\
Object Rec.           & 43.74 & 73.79 & \textbf{75.03} & \textbf{75.29} & 69.30 & 69.30 & \textbf{75.23} & 89.41\\
Positional Reas.        & 3.42 & 20.20 & 36.81 & \textbf{39.41} & 31.11 & 37.46 & \textbf{41.69} & 41.19\\
Scene Rec.            & 30.15 & 78.47 & 82.38 & 76.57 & 62.94 & 67.67 & \textbf{84.29} & 96.11\\
Sentiment Und.          & 16.50 & 73.00 & \textbf{79.50} & \textbf{82.50} & 62.50 & 28.00 & \textbf{79.50} & 60.65\\
Sport Rec.            & 28.29 & \textbf{88.45} & 88.25 & \textbf{89.84} & 77.89 & 81.27 & \textbf{89.24} & 96.20\\
Utility Affor.          & 19.88 & 66.67 & \textbf{76.02} & \textbf{74.85} & \textbf{77.19} & \textbf{73.68} & \textbf{76.02} & 21.43\\
\midrule
Micro Acc.          & 45.07 & 73.38 & 73.86 & \textbf{79.07} & 72.19 & 78.30 & \textbf{78.91} &  -\\
Macro Acc.          & 27.02 & 64.70 & 70.42 & 72.73 & 65.49 & 67.12 & \textbf{74.10} & 71.56 \\

\bottomrule
\end{tabular}
}
\end{table}

Our main results on TDIUC are detailed in Table~\ref{tab:model_performance_tdiuc}. LLaVA (13B) and LLaVA-NeXT achieve the highest micro accuracies under the asymptotic McNemar test ($p=0.2355$). GPT-4o is the next best model, showing a statistically significant difference from LLaVA (13B) ($p=0.0031$). BLIP2 obtains the poorest performance across question types, particularly in attribute/color recognition and counting. GPT-4V, GPT-4o, BLIP2, and iBLIP excel at absurd questions, whereas the LLaVA family performs worse, likely due to hallucinations. Compared to MuREl~\cite{Cadene2019}, the best system trained on TDIUC, MLLMs greatly improve for utility affordance questions, except for BLIP2.

We note that introducing absurd questions poses an additional challenge to the model. If they are not prompted to identify absurd questions, they assume every question must have a valid answer. However, when prompted to answer `does not apply,' they must consider if the question is absurd. However, this biased BLIP2 to output `does not apply' for 74.9\%  of the questions. In general, absurd questions are a test for the model's epistemic confidence in its responses.

\subsection{Assessing Counting Ability with TallyQA}

\begin{figure}[t]
\centering
\begin{subfigure}[b]{0.4\textwidth}
    \centering
    \includegraphics[width=\textwidth]{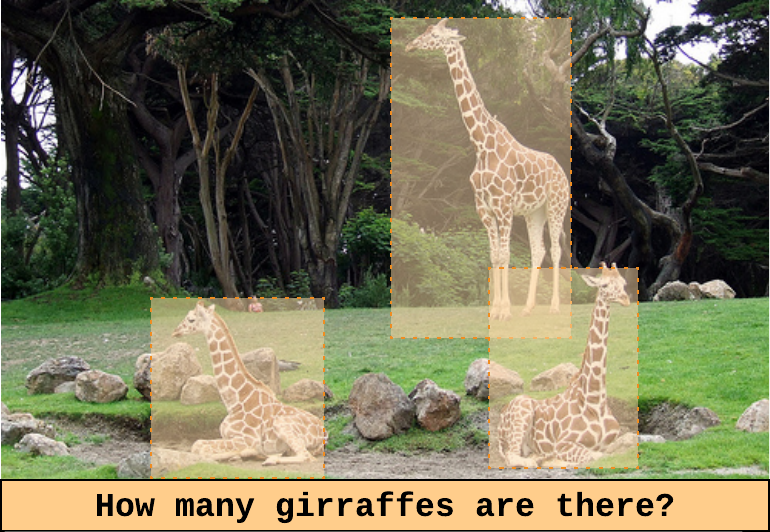}
    \caption{Simple counting question}
    \label{fig:tallyQA_simple}
\end{subfigure}
\hspace*{30pt}
\begin{subfigure}[b]{0.4\textwidth}
    \centering
    \includegraphics[width=\textwidth]{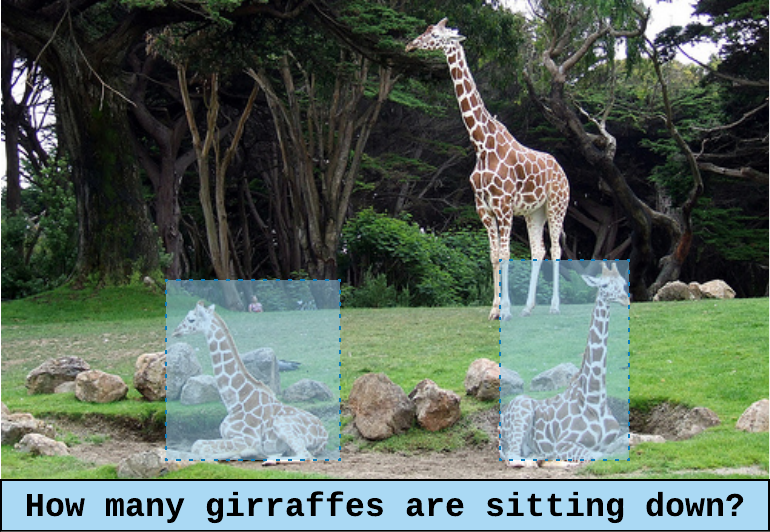}
    \caption{Complex counting question}
    \label{fig:tallyQA_complex}
\end{subfigure}
\caption{Examples of simple and complex counting questions in TallyQA.}
\label{fig:tally-qa}
\end{figure}

TallyQA~\cite{Acharya2018TallyQA} tests model's ability to count visual objects accurately. Unlike earlier VQA datasets~\cite{goyal2017making}, where the majority of the counting questions are straightforward and doable with simple object detection (e.g., ``How many giraffes are there?''), TallyQA adds additional challenges by incorporating more complex questions that necessitate detailed reasoning about the visual elements. For instance, a question such as ``How many giraffes are sitting down?'' requires the model to not only detect all the giraffes in the image but also to perform pose estimation to discern which giraffes are seated. This tests for enhanced capabilities including complex reasoning and specific visual analysis.

\paragraph{TallyQA Metrics.} In addition to reporting micro accuracy, we group the questions based on their answers (0, 1, 2, 3, or 4+) and calculate the average to determine the macro accuracy.

\paragraph{Results for TallyQA.}
\begin{table}[t]
\centering
\caption{Results on TallyQA. For Micro Acc., best performers based on paired asymptotic McNemar tests ($\alpha = 0.05$) are in bold. For Macro Acc. and RMSE, the highest value is bolded. For comparison, the result from \cite{wu2024omni} is the current best on TallyQA.} 
\label{tab:performance_tallyqa}
\begin{tabular}{lrrrrrr}
\toprule
\textbf{Model} & \multicolumn{3}{c}{\textbf{TallyQA Test-Simple}} & \multicolumn{3}{c}{\textbf{TallyQA Test-Complex}} \\
\cmidrule(lr){2-4} \cmidrule(lr){5-7}
& \textbf{Micro~Acc.} & \textbf{Macro~Acc.} & \textbf{RMSE} & \textbf{Micro~Acc.} & \textbf{Macro~Acc.} & \textbf{RMSE} \\
\midrule
BLIP2 & 64.3 & 43.0 & 3.74 & 27.5 & 24.8 & 1.57 \\
iBLIP & 73.1 & 61.7 & 1.22 & 49.3 & 35.6 & 2.15 \\
LLaVA (7B) & 75.5 & 66.5 & 1.20 & 64.1 & 45.5 & 2.21 \\
LLaVA (13B) & 76.6 & 67.3 & 1.01 & 65.6 & 47.8 & 1.93 \\
GPT-4V & 73.6 & 69.0 & 0.86 & 62.6 & 50.4 & 1.58 \\
GPT-4o & \textbf{81.5} & \textbf{74.5} & \textbf{0.60} & \textbf{71.7} & \textbf{56.9} & \textbf{1.21 }\\
LLaVA-NeXT  & 79.8 & 71.7 & 0.70 & 67.9 & 52.2 & 1.76 \\

\midrule
SMoLA~\cite{wu2024omni} & 83.3 & - & - & 70.7 & - & -\\
\bottomrule
\end{tabular}
\end{table}

The results of the TallyQA analysis are displayed in Table~\ref{tab:performance_tallyqa}. Compared to the simple counting questions, models exhibit large accuracy drops on complex counting questions, indicating deficiencies in reasoning capabilities. This is evident even for the top-performing GPT-4o, which experiences declines of 9.8\% and 17.6\% in terms of micro and macro accuracies, respectively. Additionally, as shown in Fig.~\ref{fig:tallyQA_simple_accuracy} and ~\ref{fig:tallyQA_complex_accuracy}, the accuracy of models tend to decrease as the number of objects to be counted increases, with the accuracy dropping below 30\% when the ground truth count is four or more. As shown in Figs.~\ref{fig:tallyQA_simple_accuracy} and \ref{fig:tallyQA_complex_accuracy}, the BLIP models struggled to output zero, and BLIP2 always emitted a value greater than zero.

\subsection{Assessing Chart Comprehension with DVQA}

\begin{wrapfigure}[12]{R}{0.38\textwidth}
\centering
\vspace{-57pt}
\includegraphics[width=\linewidth]{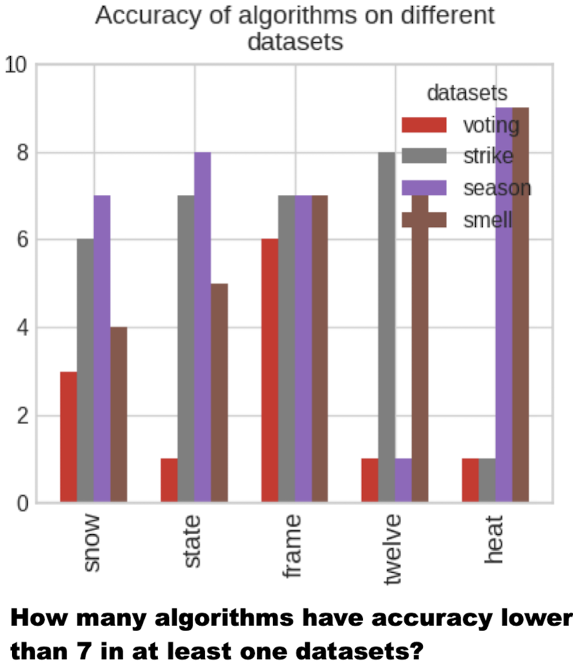}
  \caption{An example from DVQA.}
  \label{fig:dvqa-dataset}
\end{wrapfigure}

DVQA~\cite{kafle2018dvqa} is a VQA dataset evaluating chart understanding. DVQA requires the model to perform grounding extensively. With synthetic charts, the model is required to handle words or formulae that are specific for that instance. This contrasts with datasets using natural images, where questions such as ``What color is the sky?'' are based on universal concepts, and even models that simply exploit dataset biases can obtain high accuracy by guessing that the sky is either blue or gray. In contrast, the models cannot inflate accuracy by exploiting such correlations in DVQA since the concepts correspond to arbitrary values (e.g., the labels can correspond to arbitrary bar heights and colors)~\cite{kafle2018dvqa}.

\vspace{20pt}

\textbf{DVQA Metrics.} For DVQA, we report micro and macro accuracy. DVQA has 3 question types: structural understanding, data retrieval, and reasoning. They are averaged to compute macro accuracy. 

\begin{table}[t]
\centering
\caption{Percentage (\%) accuracy results on DVQA. Best performers based on paired asymptotic McNemar tests ($\alpha = 0.05$) are in bold, except for Macro. Acc., where the max is bolded. For comparison, PReFIL and Human results correspond to performance on Test-Novel~\cite{kafle2020answering}, where PReFIL uses Improved OCR (see \cite{kafle2020answering}). PReFIL is a DVQA system trained on DVQA's training set.}
\label{tab:dvqa-results}
\begin{tabular}{@{}lccc|cc@{}}
\toprule
\textbf{Model} & \textbf{Reasoning} & \textbf{Retrieval} & \textbf{Structural} & \textbf{Micro Acc.} & \textbf{Macro Acc.}\\ \midrule
BLIP2 & 12.79 & 9.38 & 45.78 & 16.17 & 22.65\\
iBLIP & 15.22 & 14.23 & 48.50 & 19.41 & 25.98\\
LLaVA~(7B) & 17.76 & 20.22 & 51.40 & 23.10 & 29.79\\
LLaVA~(13B) & 19.01 & 22.07 & 57.89 & 25.25 & 32.99\\
GPT-4V & 33.26 & 61.83 & 88.73 & 49.88 & 61.27\\
GPT-4o & 52.06 & 73.64 & \textbf{95.60} & 64.84 & 73.77\\ 
LLaVA-NeXT & \textbf{69.14} & \textbf{82.73} & 73.47 & \textbf{74.06} & \textbf{75.11}\\
\midrule
PReFIL~\cite{kafle2020answering} & 80.73 & 67.13 & 99.57 & 80.04 & -\\
Human~\cite{kafle2020answering} & 85.83 & 88.70 & 96.19 & 88.18 & -\\
\bottomrule
\end{tabular}

\end{table}

\paragraph{Results for DVQA.}
Results for DVQA are given in Table~\ref{tab:dvqa-results}. LLaVA-NeXT achieved the highest micro accuracy, and under an asymptotic McNemar test all other models had a statistically significant difference in micro accuracy ($p<0.0001$). Compared to other categories, all models performed best on structural questions. Structural questions include questions such as: 1) ``How many bars are there?'' 2) ``Does the chart contain any negative values?'' 3) ``Are the bars horizontal?'' and 4) ``Is each bar a single solid color without patterns?'' These questions do not require extracting textual information from the image and only require the analysis of visual features. Models were worst at reasoning questions. Our results highlight the importance of training on synthetic data, as was done in LLaVA-NeXT, for achieving strong performance. No MLLM achieves the performance of a PReFIL for reasoning questions, which was trained on DVQA's training set, or of humans~\cite{kafle2020answering}.


\subsection{Analyzing the Strengths and Weaknesses of Today's MLLMs}
 Generally, we found that all evaluated models perform poorly in detecting multiple objects in referring expressions. The results from TallyQA align with this observation, showing a performance decline as the counting number increases. Furthermore, introducing absurd questions and allowing models to answer `no' presents an additional challenge. The models must be more confident in their selections; otherwise, they risk misclassifying answerable questions as absurd, leading to a significant performance drop. The results from DVQA indicate that the challenges posed by natural scene datasets are very different from those in synthetic image datasets. While some models demonstrate robustness to DVQA, others, such as LLaVA, BLIP2, and iBLIP, exhibit a significant performance gap when evaluated on natural versus synthetic image datasets.

\section{Related Work}

\paragraph{Problems with Widely Used Datasets.}
With the advent of large foundation models, datasets for training, fine-tuning, and validation have become increasingly important~\cite{liang2022holistic}. These datasets are pivotal in reflecting a model's performance across different aspects. Notably, many recent MLLMs rely on some of the earliest established datasets~\cite{goyal2017making,refcocokazemzadeh2014referitgame,cocoqa}, which, while foundational, are increasingly recognized for their constraints and biases. Existing VQA datasets have several well-known issues. Most fail to properly assess grounding capabilities—linking specific parts of an image to corresponding textual elements in questions. For example, on some datasets, models can achieve approximately 50\% accuracy even when blinded to the image, relying solely on the questions~\cite{Kafle2016AnswerType}. This indicates that many questions do not depend on grounding capabilities, allowing models to exploit learned biases rather than visual evidence.
Moreover, popular VQA datasets focus narrowly on specific question types, limiting the assessment of models' generalization abilities. Most questions (69.84\%) ask about objects in the image, hindering the model's ability to handle abstract reasoning, complex visual cues, or nuanced human interactions. Additionally, MLLMs often are not evaluated on synthetic datasets, missing opportunities to reveal limitations not observed with natural images. 
Mainstream referring expression recognition datasets like RefCOCO typically assume each referring expression refers to a single object, oversimplifying the task. In RefCOCOg~\cite{refcocogmao2016}, it was shown~\cite{cirik2018visual} that randomly permuting words in the referring expressions only reduced performance by 5\%, and models could achieve 71.2\% precision for the top-2 predictions using only the image. This suggests that models exploit dataset quirks and biases rather than utilizing linguistic cues for grounding. The imbalance in target object selection and the simplistic design of referring expressions, with only one associated bounding box, further exacerbate this issue.

\paragraph{Related Efforts to Improve MLLM Evaluation.}
Recent works highlight challenges in evaluating MLLMs. In \cite{yuksekgonul2022when}, the ARO benchmark was introduced to assess models' understanding of complex compositional elements, and models evaluated on it performed poorly for like ``the grass is eating the horse'' versus ``the horse is eating grass.'' Similarly, the Winoground datasets~\cite{thrush2022winoground} require models to match images with captions that use identical words in different orders to assess their comprehension of linguistic composition concerning visual information. In~\cite{shah2019cycle}, a cycle-consistency framework is proposed, evaluating models' ability to understand semantically similar questions. These studies complement ours and reveal other biases and limitations in MLLMs.

\section{Discussion}

Our TallyQA results highlight the necessity of incorporating more complex counting questions to reflect models' counting capabilities better. The LLaVA family demonstrates robustness to complex counting questions that demand sophisticated reasoning. In contrast, other models, like BLIP2, perform poorly on these complex questions despite performing adequately on easy counting questions compared to LLaVA. Relying solely on easy counting questions can lead to inflated scores, which can be misleading.

Results from VQDv1 show that traditional single-object referring expressions are more accessible for models to handle. However, introducing more targets in referring expressions presents a significant challenge, as performance drops when more objects are involved. Examining VQDv1 and TallyQA, they are complementary in evaluating models. In VQDv1, the model must generate one or more bounding boxes around objects described in the question, serving as an improved version of counting questions by requiring models to justify their answers. In TallyQA, models perform well when accounting for fewer objects, but performance drops significantly as the number of objects increases, indicating poor generalization abilities. This aligns with findings from VQDv1, where models struggle with multiple bounding boxes but perform well with a single bounding box.  VQDv1 and TallyQA offer a comprehensive evaluation of a model’s ability to justify its answers and handle varying numbers of objects, highlighting weaknesses in object detection and counting abilities.

Results from TDIUC provide insight into models' generalization across different question types. Most perform poorly on positional reasoning, an essential skill for complex counting questions and referring expressions. TDIUC also includes counting questions, and similar to TallyQA, models show a significant drop in macro accuracy. However, these results also show that Utility/Affordance questions benefit greatly from MLLMs compared to models trained on TDIUC. 

All models perform poorly on DVQA, indicating that MLLMs struggle with parsing chart information, especially in reasoning and data retrieval questions. LLaVA-NeXT improves significantly over other open-source MLLMs on DVQA, likely due to its training on documents and diagrams. The DVQA dataset highlights the challenges presented by synthetic images.

\textbf{Societal Implications.} MLLMs are increasingly integrated into various applications, from virtual assistants to automated content generation. However, current popular datasets often fail to capture models' limitations. Our study informs where MLLMs can be safely used and identifies tasks they are not yet ready for, emphasizing the importance of rigorous evaluation datasets.

\textbf{Limitations.} One significant challenge we encountered was effectively prompting the models (see Appendices~\ref{sec:prompt-selection} and \ref{sec:vqd-appendix}). The performance of MLLMs is susceptible to the phrasing and structure of prompts, with small changes leading to significant variations in outputs. Crafting prompts that balance complexity and clarity is difficult, especially given the diversity of tasks and datasets. Additionally, no standardized approach to prompt engineering across different models complicates fair comparisons. We experimented with various formulations to find effective prompts, but our approach may still have limitations. Future work should focus on developing systematic and standardized methods for prompt engineering to ensure consistent and fair evaluations.

\section{Conclusions}
In this paper, we conducted comprehensive, skill-specific evaluations of MLLMs released in 2023--2024. Our analysis revealed several weaknesses that are not apparent when using mainstream datasets alone. To enhance accessibility for researchers and facilitate benchmark comparisons, we have integrated these datasets into a fork of the widely used LAVIS framework~\cite{li-etal-2023-lavis}, and we will work with the LAVIS team to merge our version into the main trunk or release it as a separate entity, if necessary.

\begin{ack}
This work was supported in part by NSF award \#2326491. The views and conclusions contained herein are those of the authors and should not be interpreted as representing the official policies or endorsements of any sponsor. Figures \ref{fig:vqd-dataset}, \ref{fig:tdiuc}, \ref{fig:tally-qa}, and \ref{fig:dvqa-dataset} are reproduced from their original papers, which were authored by members of our team.
\end{ack}

\bibliographystyle{unsrtnat}
{
\small
\bibliography{library}

\begin{thebibliography}{35}
\providecommand{\natexlab}[1]{#1}
\providecommand{\url}[1]{\texttt{#1}}
\expandafter\ifx\csname urlstyle\endcsname\relax
  \providecommand{\doi}[1]{doi: #1}\else
  \providecommand{\doi}{doi: \begingroup \urlstyle{rm}\Url}\fi

\bibitem[Li et~al.(2023{\natexlab{a}})Li, Li, Savarese, and Hoi]{li2023blip}
Junnan Li, Dongxu Li, Silvio Savarese, and Steven Hoi.
\newblock Blip-2: Bootstrapping language-image pre-training with frozen image encoders and large language models.
\newblock \emph{arXiv preprint arXiv:2301.12597}, 2023{\natexlab{a}}.

\bibitem[Chowdhery et~al.(2022)Chowdhery, Narang, Devlin, Bosma, Mishra, Roberts, Barham, Chung, Sutton, Gehrmann, et~al.]{chowdhery2022palm}
Aakanksha Chowdhery, Sharan Narang, Jacob Devlin, Maarten Bosma, Gaurav Mishra, Adam Roberts, Paul Barham, Hyung~Won Chung, Charles Sutton, Sebastian Gehrmann, et~al.
\newblock Palm: Scaling language modeling with pathways.
\newblock \emph{arXiv preprint arXiv:2204.02311}, 2022.

\bibitem[Zhu et~al.(2023)Zhu, Chen, Shen, Li, and Elhoseiny]{zhu2023minigpt}
Deyao Zhu, Jun Chen, Xiaoqian Shen, Xiang Li, and Mohamed Elhoseiny.
\newblock Minigpt-4: Enhancing vision-language understanding with advanced large language models.
\newblock \emph{arXiv preprint arXiv:2304.10592}, 2023.

\bibitem[Koh et~al.(2023)Koh, Salakhutdinov, and Fried]{koh2023grounding}
Jing~Yu Koh, Ruslan Salakhutdinov, and Daniel Fried.
\newblock Grounding language models to images for multimodal inputs and outputs.
\newblock \emph{ICML}, 2023.

\bibitem[Liu et~al.(2023{\natexlab{a}})Liu, Li, Wu, and Lee]{liu2023llava}
Haotian Liu, Chunyuan Li, Qingyang Wu, and Yong~Jae Lee.
\newblock Visual instruction tuning.
\newblock In \emph{NeurIPS}, 2023{\natexlab{a}}.

\bibitem[Liu et~al.(2023{\natexlab{b}})Liu, Li, Li, and Lee]{liu2023improved}
Haotian Liu, Chunyuan Li, Yuheng Li, and Yong~Jae Lee.
\newblock Improved baselines with visual instruction tuning.
\newblock \emph{arXiv preprint arXiv:2310.03744}, 2023{\natexlab{b}}.

\bibitem[Goyal et~al.(2017)]{goyal2017making}
Yash Goyal et~al.
\newblock Making the v in vqa matter: Elevating the role of image understanding in visual question answering.
\newblock In \emph{2017 IEEE Conference on Computer Vision and Pattern Recognition (CVPR)}, pages 6325--6333. IEEE, July 2017.
\newblock \doi{10.1109/CVPR.2017.670}.

\bibitem[Zhang et~al.(2016)]{zhang2016yin}
Peng Zhang et~al.
\newblock Yin and yang: Balancing and answering binary visual questions.
\newblock In \emph{2016 IEEE Conference on Computer Vision and Pattern Recognition (CVPR)}. IEEE, June 2016.
\newblock \doi{10.1109/CVPR.2016.542}.

\bibitem[Sharma et~al.(2018)Sharma, Ding, Goodman, and Soricut]{sharma2018conceptual}
Piyush Sharma, Nan Ding, Sebastian Goodman, and Radu Soricut.
\newblock Conceptual captions: A cleaned, hypernymed, image alt-text dataset for automatic image captioning.
\newblock In \emph{Proceedings of the 56th Annual Meeting of the Association for Computational Linguistics (Volume 1: Long Papers)}, pages 2556--2565, 2018.

\bibitem[sha()]{sharegpt}
{S}hare{G}{P}{T}: {S}hare your wildest {C}hat{G}{P}{T} conversations with one click. --- sharegpt.com.
\newblock \url{https://sharegpt.com/}.
\newblock [Accessed 16-05-2024].

\bibitem[Ren et~al.(2015)Ren, Kiros, and Zemel]{cocoqa}
Menglin Ren, Ryan Kiros, and Richard Zemel.
\newblock Exploring models and data for image question answering.
\newblock In \emph{Advances in Neural Information Processing Systems}, 2015.

\bibitem[Kazemzadeh et~al.(2014)]{refcocokazemzadeh2014referitgame}
Sahar Kazemzadeh et~al.
\newblock Referitgame: Referring to objects in photographs of natural scenes.
\newblock In \emph{Proceedings of the 2014 Conference on Empirical Methods in Natural Language Processing (EMNLP)}, 2014.
\newblock \doi{10.3115/v1/d14-1086}.

\bibitem[Acharya et~al.(2019)Acharya, Jariwala, et~al.]{Acharya2019VQD}
Manoj Acharya, Karan Jariwala, et~al.
\newblock Vqd: Visual query detection in natural scenes.
\newblock \emph{arXiv preprint arXiv:1904.02794}, 2019.
\newblock URL \url{https://arxiv.org/abs/1904.02794}.

\bibitem[Acharya et~al.(2018)Acharya, Kafle, et~al.]{Acharya2018TallyQA}
Manoj Acharya, Kushal Kafle, et~al.
\newblock Tallyqa: Answering complex counting questions.
\newblock \emph{arXiv preprint arXiv:1810.12440}, 2018.
\newblock URL \url{https://arxiv.org/abs/1810.12440}.

\bibitem[Kafle and Kanan(2017)]{kafle2017analysis}
Kushal Kafle and Christopher Kanan.
\newblock An analysis of visual question answering algorithms.
\newblock In \emph{ICCV}, 2017.

\bibitem[Kafle et~al.(2018)Kafle, Price, Cohen, and Kanan]{kafle2018dvqa}
Kushal Kafle, Brian Price, Scott Cohen, and Christopher Kanan.
\newblock Dvqa: Understanding data visualizations via question answering.
\newblock In \emph{Proceedings of the IEEE conference on computer vision and pattern recognition}, pages 5648--5656, 2018.

\bibitem[gpt(2024)]{gpt4o}
Hello gpt-4o.
\newblock \url{https://openai.com/index/hello-gpt-4o/}, 2024.
\newblock [Accessed 16-05-2024].

\bibitem[Mao et~al.(2016)]{refcocogmao2016}
Junhua Mao et~al.
\newblock Generation and comprehension of unambiguous object descriptions.
\newblock In \emph{2016 IEEE Conference on Computer Vision and Pattern Recognition (CVPR)}. IEEE, June 2016.
\newblock \doi{10.1109/CVPR.2016.9}.
\newblock URL \url{https://doi.org/10.1109/cvpr.2016.9}.

\bibitem[Kafle and Kanan(2016)]{Kafle2016AnswerType}
Kushal Kafle and Christopher Kanan.
\newblock Answer-type prediction for visual question answering.
\newblock In \emph{2016 IEEE Conference on Computer Vision and Pattern Recognition (CVPR)}, June 2016.
\newblock \doi{10.1109/CVPR.2016.538}.
\newblock URL \url{https://doi.org/10.1109/cvpr.2016.538}.

\bibitem[Cirik et~al.(2018)]{cirik2018visual}
Volkan Cirik et~al.
\newblock Visual referring expression recognition: What do systems actually learn?
\newblock In \emph{Proceedings of the 2018 Conference of the North American Chapter of the Association for Computational Linguistics: Human Language Technologies, Volume 2 (Short Papers)}, 2018.
\newblock \doi{10.18653/v1/n18-2123}.
\newblock URL \url{https://doi.org/10.18653/v1/n18-2123}.

\bibitem[Akula et~al.(2020)Akula, Gella, Al-Onaizan, Zhu, and Reddy]{akula2020words}
Arjun~R Akula, Spandana Gella, Yaser Al-Onaizan, Song-Chun Zhu, and Siva Reddy.
\newblock Words aren't enough, their order matters: On the robustness of grounding visual referring expressions.
\newblock \emph{arXiv preprint arXiv:2005.01655}, 2020.

\bibitem[Liu et~al.(2024{\natexlab{a}})Liu, Li, Wu, and Lee]{liu2024visual}
Haotian Liu, Chunyuan Li, Qingyang Wu, and Yong~Jae Lee.
\newblock Visual instruction tuning.
\newblock \emph{Advances in neural information processing systems}, 36, 2024{\natexlab{a}}.

\bibitem[Dai et~al.(2024)Dai, Li, Li, Tiong, Zhao, Wang, Li, Fung, and Hoi]{dai2024instructblip}
Wenliang Dai, Junnan Li, Dongxu Li, Anthony Meng~Huat Tiong, Junqi Zhao, Weisheng Wang, Boyang Li, Pascale~N Fung, and Steven Hoi.
\newblock Instructblip: Towards general-purpose vision-language models with instruction tuning.
\newblock \emph{Advances in Neural Information Processing Systems}, 36, 2024.

\bibitem[Liu et~al.(2024{\natexlab{b}})Liu, Li, Li, Li, Zhang, Shen, and Lee]{liu2024llavanext}
Haotian Liu, Chunyuan Li, Yuheng Li, Bo~Li, Yuanhan Zhang, Sheng Shen, and Yong~Jae Lee.
\newblock Llava-next: Improved reasoning, ocr, and world knowledge, January 2024{\natexlab{b}}.
\newblock URL \url{https://llava-vl.github.io/blog/2024-01-30-llava-next/}.

\bibitem[Achiam et~al.(2023)Achiam, Adler, Agarwal, Ahmad, Akkaya, Aleman, Almeida, Altenschmidt, Altman, Anadkat, et~al.]{achiam2023gpt}
Josh Achiam, Steven Adler, Sandhini Agarwal, Lama Ahmad, Ilge Akkaya, Florencia~Leoni Aleman, Diogo Almeida, Janko Altenschmidt, Sam Altman, Shyamal Anadkat, et~al.
\newblock Gpt-4 technical report.
\newblock \emph{arXiv preprint arXiv:2303.08774}, 2023.

\bibitem[Yang et~al.(2023)Yang, Li, Lin, Wang, Lin, Liu, and Wang]{yang2023dawn}
Zhengyuan Yang, Linjie Li, Kevin Lin, Jianfeng Wang, Chung-Ching Lin, Zicheng Liu, and Lijuan Wang.
\newblock The dawn of lmms: Preliminary explorations with gpt-4v (ision).
\newblock \emph{arXiv preprint arXiv:2309.17421}, 9\penalty0 (1):\penalty0 1, 2023.

\bibitem[Cadene et~al.(2019)Cadene, Ben-younes, Cord, and Thome]{Cadene2019}
Remi Cadene, Hamid Ben-younes, Matthieu Cord, and Nicolas Thome.
\newblock Murel: Multimodal relational reasoning for visual question answering.
\newblock \emph{arXiv.org}, February 2019.
\newblock URL \url{https://arxiv.org/abs/1902.09487}.

\bibitem[Wu et~al.(2024)Wu, Hu, Wang, Pang, and Soricut]{wu2024omni}
Junnan Wu, Xun Hu, Yongyi Wang, Bo~Pang, and Radu Soricut.
\newblock Omni-smola: Boosting generalist multimodal models with soft mixture of low-rank experts.
\newblock \emph{arXiv.org}, April 2 2024.
\newblock \url{https://arxiv.org/abs/2312.00968}.

\bibitem[Kafle et~al.(2020)Kafle, Shrestha, Cohen, Price, and Kanan]{kafle2020answering}
Kushal Kafle, Robik Shrestha, Scott Cohen, Brian Price, and Christopher Kanan.
\newblock Answering questions about data visualizations using efficient bimodal fusion.
\newblock In \emph{Proceedings of the IEEE/CVF Winter conference on applications of computer vision}, pages 1498--1507, 2020.

\bibitem[Liang et~al.(2022)Liang, Bommasani, Lee, Tsipras, Soylu, Yasunaga, Zhang, Narayanan, Wu, Kumar, et~al.]{liang2022holistic}
Percy Liang, Rishi Bommasani, Tony Lee, Dimitris Tsipras, Dilara Soylu, Michihiro Yasunaga, Yian Zhang, Deepak Narayanan, Yuhuai Wu, Ananya Kumar, et~al.
\newblock Holistic evaluation of language models.
\newblock \emph{arXiv preprint arXiv:2211.09110}, 2022.

\bibitem[Yuksekgonul et~al.(2022)Yuksekgonul, Bianchi, Kalluri, Jurafsky, and Zou]{yuksekgonul2022when}
Mert Yuksekgonul, Federico Bianchi, Pratyusha Kalluri, Dan Jurafsky, and James Zou.
\newblock When and why vision-language models behave like bags-of-words, and what to do about it.
\newblock \emph{OpenReview}, Sep 2022.
\newblock Available at \url{https://openreview.net/forum?id=KRLUvxh8uaX}.

\bibitem[Thrush et~al.(2022)Thrush, Jiang, Bartolo, Singh, Williams, Kiela, and Ross]{thrush2022winoground}
Tristan Thrush, Ryan Jiang, Max Bartolo, Amanpreet Singh, Adina Williams, Douwe Kiela, and Candace Ross.
\newblock Winoground: Probing vision and language models for visio-linguistic compositionality, Apr 2022.
\newblock Available at \url{https://arxiv.org/abs/2204.03162}.

\bibitem[Shah et~al.(2019)Shah, Chen, Rohrbach, and Parikh]{shah2019cycle}
Meet Shah, Xinlei Chen, Marcus Rohrbach, and Devi Parikh.
\newblock Cycle-consistency for robust visual question answering, Feb 2019.
\newblock Available at \url{https://arxiv.org/abs/1902.05660}.

\bibitem[Li et~al.(2023{\natexlab{b}})Li, Li, Le, Wang, Savarese, and Hoi]{li-etal-2023-lavis}
Dongxu Li, Junnan Li, Hung Le, Guangsen Wang, Silvio Savarese, and Steven~C.H. Hoi.
\newblock {LAVIS}: A one-stop library for language-vision intelligence.
\newblock In \emph{Proceedings of the 61st Annual Meeting of the Association for Computational Linguistics (Volume 3: System Demonstrations)}, pages 31--41, Toronto, Canada, July 2023{\natexlab{b}}. Association for Computational Linguistics.
\newblock URL \url{https://aclanthology.org/2023.acl-demo.3}.

\bibitem[University(2010)]{WordNet}
Princeton University.
\newblock \emph{WordNet: An Electronic Lexical Database}.
\newblock Princeton University, 2010.
\newblock \url{https://wordnet.princeton.edu}.

\end{thebibliography}
}

\clearpage

\appendix

\begin{center}
    {\Large{\textbf{Appendix}}}
\end{center}

\section{Computational Resources}
The evaluations of open-source MLLMs were conducted on a single A100 GPU with 40GB of RAM, which required approximately 200 hours on our university-wide computing infrastructure. To evaluate GPT-4V/GPT-4o, which are closed-source, we used the paid ChatGPT API provided by OpenAI and spent \$922 for GPT-4V and \$451 for GPT-4o, including runs to tune prompts.

\section{Additional Dataset Details}
In this section, we present additional dataset details.

\subsection{TallyQA}
The counting questions in TallyQA are classified into complex and simple counting questions~\cite{Acharya2018TallyQA}. Simple counting questions were imported from existing datasets like VQA2 and Visual Genome. Complex questions were collected using Amazon Mechanical Turk (AMT) to gather 19,500 complex questions for 17,545 unique images. The images were sourced from both COCO and Visual Genome to ensure variety.  The testing set of TallyQA contains 38,589 questions, which is a reasonable size. Therefore, we evaluated models on the entire original test set. The distribution of unique answers is given in Table~\ref{tab:TallyQA_Distribution}. TallyQA is provided under the terms of the Apache License Version 2.0, January 2004: \url{http://www.apache.org/licenses/}
\begin{table}[h]
\centering
\caption{The distribution of unique answers in TallyQA.}
\label{tab:TallyQA_Distribution}
\begin{tabular}{lrr}
\toprule
\textbf{Answer}  & \textbf{Complex} & \textbf{Simple}\\ \midrule
zero & 4335 & 637\\
one & 6853 & 12308\\
two & 2479 & 5636\\
three & 901 & 2034\\
four & 453 & 1101\\
five & 195 & 435\\
six & 133 & 319\\
seven & 70 & 152\\
eight & 69 & 145\\
nine & 31 & 84\\
ten & 33 & 48\\
eleven & 12 & 30\\
twelve & 25 & 33\\
thirteen & 7 & 13\\
fourteen & 6 & 9\\
fifteen & 6 & 7\\
\bottomrule
\end{tabular}
\end{table}

\subsection{VQDv1}
VQDv1~\cite{Acharya2019VQD} was created synthetically using annotations from Visual Genome, COCO, and COCO Panoptic. This synthetic generation approach helps combat certain biases. The queries are generated using multiple templates for each type, allowing for diverse queries. The annotations used to generate these questions are derived from a combination of COCO's object annotations and Visual Genome's attribute and relationship information.

For VQDv1, almost 90\% of the queries have less than two ground truth bounding boxes. In our subset, we retained all queries with more than one ground truth bounding box, and we sampled 10\% of the queries with zero or one ground truth bounding box. Table~\ref{tab:bounding_boxes} provides the distribution of ground truth boxes across queries. The VQDv1 dataset is provided under the terms of the Creative Commons Attribution 4.0 International (CC BY 4.0) license: \url{https://creativecommons.org/licenses/by/4.0/legalcode} 
\begin{table}[t]
    \centering
    \caption{Bounding box distribution for the original and modified versions of VQDv1.}
    \label{tab:bounding_boxes}
    \begin{tabular}{crr}
        \hline
        \textbf{Bounding Box Count} & \textbf{Original Version} & \textbf{Our Version} \\
        \hline
        0  & 80025 (42.08\%) & 8001 (21.59\%) \\
        1  & 90101 (47.38\%) & 9008 (24.31\%) \\
        2  & 10127 ~~(5.33\%)  & 10127 (27.33\%) \\
        3  & 3200 ~~(1.68\%)   & 3200 ~~(8.64\%)  \\
        4  & 1894 ~~(1.00\%)   & 1894 ~~(5.11\%)  \\
        5  & 1334 ~~(0.70\%)   & 1334 ~~(3.60\%)  \\
        6  & 700 ~~(0.37\%)    & 700 ~~(1.89\%)   \\
        7  & 533 ~~(0.28\%)    & 533 ~~(1.44\%)   \\
        8  & 366 ~~(0.19\%)    & 366 ~~(0.99\%)   \\
        9  & 305 ~~(0.16\%)    & 305 ~~(0.82\%)   \\
        10 & 276 ~~(0.15\%)    & 276 ~~(0.74\%)   \\
        11 & 193 ~~(0.10\%)    & 193 ~~(0.52\%)   \\
        12 & 194 ~~(0.10\%)    & 194 ~~(0.52\%)   \\
        13 & 255 ~~(0.13\%)    & 255 ~~(0.69\%)   \\
        14 & 618 ~~(0.32\%)    & 618 ~~(1.67\%)   \\
        15 & 26 ~~(0.01\%)     & 26 ~~(0.07\%)    \\
        16 & 5 ~~(0.00\%)      & 5 ~~(0.01\%)     \\
        17 & 7 ~~(0.00\%)      & 7 ~~(0.02\%)     \\
        18 & 7 ~~(0.00\%)      & 7 ~~(0.02\%)     \\
        19 & 3 ~~(0.00\%)      & 3 ~~(0.01\%)     \\
        20 & 1 ~~(0.00\%)      & 1 ~~(0.00\%)     \\
        23 & 2 ~~(0.00\%)      & 2 ~~(0.01\%)     \\
        25 & 1 ~~(0.00\%)      & 1 ~~(0.00\%)     \\
        26 & 1 ~~(0.00\%)      & 1 ~~(0.00\%)     \\
        \hline
    \end{tabular}    
\end{table}

\subsection{DVQA}
The DVQA dataset was created by synthetically generating bar charts to test multiple aspects of bar chart understanding. This automatic generation process allows precise control over the visual elements' positions and appearances, and provides access to meta-data about the elements in the image, which is not available with real data~\cite{kafle2018dvqa}.

The original version of DVQA had two test sets: Test-Familiar and Test-Novel. The critical difference between these two sets is that every bar chart in Test-Familiar has labels in DVQA's training set, whereas Test-Novel does not. Given that we are conducting zero-shot evaluations, these two sets can be treated equivalently. Therefore, we sample the same number of questions from both. Table~\ref{tab:dvqa_distribution} shows the question distributions of our subset version of DVQA. The DVQA dataset is provided under the terms of the Creative Commons Attribution 4.0 International (CC BY 4.0): \url{https://creativecommons.org/licenses/by/4.0/legalcode}
\begin{table}[H]
\centering
\caption{Distribution of question types in DVQA.}
\label{tab:dvqa_distribution}
\begin{tabular}{lrrr}
\toprule
\textbf{Question Type} & \textbf{Test-Familiar Version} & \textbf{Test-Novel Version} & \textbf{Our Version} \\
\midrule
Data        & 185356 (31.93\%) & 185452 (31.90\%)  & 9269 (31.91\%)\\
Reasoning   & 316923 (54.59\%) & 316881 (54.51\%) & 15844 (54.55\%) \\
Structure   & 78278 (13.48\%)  & 78988 (13.59\%)  & 3930 (13.53\%)\\
\bottomrule
\end{tabular}
\end{table}

\subsection{TDIUC}
The TDIUC dataset was created by incorporating questions from three sources: existing datasets, questions generated based on image annotations, and human annotators. Questions were imported from COCO-VQA and Visual Genome datasets, with templates and regular expressions used to classify and generate questions\cite{kafle2017analysis}. Additionally, questions were generated using COCO’s semantic segmentation annotations and Visual Genome’s objects and attribute annotations\cite{kafle2017analysis}. For certain question types like sentiment understanding and object utility/affordance, trained volunteers performed manual annotation using a web-based tool\cite{kafle2017analysis}. We sample proportionately from 12 question types in TDIUC. Table ~\ref{tab:tdiuc_distribution} shows our subset of TDIUC question distributions. TDIUC is a public dataset but does not mention a particular license.\url{https://kushalkafle.com/projects/tdiuc.html}
\begin{table}[H]
\centering
\caption{Distribution of question types in TDIUC.}
\label{tab:tdiuc_distribution}
\begin{tabular}{lrr}
\toprule
\textbf{Question Type} &  \textbf{Original Version}  & \textbf{Our Version}\\
\midrule
Absurd                 & 120411 (22.35\%) & 6844 (25.00\%) \\
Activity Recognition   & 2682 ~~(0.50\%) & 77 ~~(0.28\%)  \\
Attribute              & 9200 ~~(1.71\%)  & 296 ~~(1.08\%)  \\
Color                  & 62490 (11.60\%) & 2142 ~~(7.82\%)  \\
Counting               & 52905 ~~(9.82\%)  & 2262 ~~(8.26\%)  \\
Object Presence        & 215324 (39.96\%) & 11884 (43.41\%) \\
Object Recognition     & 30693 ~~(5.70\%)  & 1646 ~~(6.01\%)  \\
Positional Reasoning   & 12284 ~~(2.28\%)  & 523 ~~(1.91\%)  \\
Scene Recognition      & 22032 ~~(4.09\%)  & 1188 ~~(4.34\%)  \\
Sentiment Understanding& 634 ~~(0.12\%)  & 27 ~~(0.10\%)  \\
Sport Recognition      & 10042 ~~(1.86\%)  & 478 ~~(1.75\%)  \\
Utility Affordance     & 171 ~~(0.03\%)  & 12 ~~(0.04\%)  \\
\bottomrule
\end{tabular}
\end{table}

\section{Prompt Engineering}
\label{sec:prompt-selection}

To make the model performance comparison as fair as possible, we endeavored to keep the prompts consistent across different models. However, this was challenging due to variations in the models' ability to process the prompts. For example, BLIP2 and iBLIP failed when prompted to answer using a template such as ``My answer is <answer>.''  Inspired by \citet{liu2023improved}, for TDIUC, DVQA, and TallyQA, we prompt the models to answer as concisely as possible instead of asking them to generate entire sentences. These prompts are given in Fig.~\ref{fig:instruction_templates}.

\begin{figure}[H]
\centering
\caption{Prompts used for TallyQA, DVQA, and TDIUC.}
\label{fig:instruction_templates}
\vspace{-10pt}
\begin{tcolorbox}[colback=gray!10, colframe=black, width=\textwidth, boxrule=0.5mm, arc=2mm, auto outer arc]

\begin{itemize}[leftmargin=*]
    \item TallyQA: Please answer the question in one word.
    \item DVQA: Please answer the question in one word
    \item TDIUC: Please answer in one word. Answer `doesnotapply' if the question is not related to the image or cannot be answered.
\end{itemize}
\end{tcolorbox}

\end{figure}

Despite much effort, for VQDv1, we were unable to identify a universal prompt for generating multiple bounding boxes that worked well across models. For example, as shown in Table.~\ref{tab:llava-vqd-struggle}, LLaVA (7B) repeatedly generated the same bounding boxes until the maximum token limit was reached when this prompt was used. We believe this occurs because the model is confused by the instruction to generate multiple bounding boxes, even when only one object is detected. This may explain why it repeatedly generates the same bounding box. While we considered non-maximal suppression or eliminating redundant boxes, our goal is to fairly evaluate MLLMs without excessively post-processing their outputs. Therefore, we fine-tuned the prompts for different models. The results reported in the paper represent the best outcomes from our evaluations. The best-identified prompts for each model on VQDv1 are given in Table~\ref{tab:vqd-prompts}.

\begin{table}[H]
    \centering
    \caption{LLaVA (7B) struggled with some prompts for VQDv1.}
    \label{tab:llava-vqd-struggle}
    \begin{tabular}{|>{\raggedright\arraybackslash}p{3cm}|>{\raggedright\arraybackslash}p{10cm}|}
        \hline
    \textbf{Prompt with Query} & Where is the motorcycle? Instruction: \textbf{Generate a list of bounding box coordinates around the objects that the prompt mentioned if they exist in the image. Even if the question uses a singular verb like `is', you should still generate multiple bounding boxes if there are multiple objects that satisfy the prompt. The bounding box list should be in the following format: [[x\_min, y\_min, x\_max, y\_max], [x\_min, y\_min, x\_max, y\_max]]} \\
        \hline
        \textbf{LLaVA (7B)} & predicted\_answer: ``[[0.01, 0.0, 0.28, 0.24], [0.01, 0.0, 0.28, 0.24], [0.01, 0.0, 0.28, 0.24], [0.01, 0.0, 0.28, 0.24], [0.01, 0.0, 0.28, 0.24], [0.01, 0.0, 0.28, 0.24], [0.01, 0.0, 0.28, 0.24], [0.01, 0.0, 0.28, 0.24], [0.01, 0.0, 0.28, 0.24], [0.01, 0.0, 0.28, 0.24], [0.01, 0.0, 0.28, 0.24], [0.01, 0.0, 0.28, 0.24], [0.01, 0.0, 0.28, 0.24], [0.01, 0.0, 0.28, 0.24], [0.01, 0.0, 0.28, 0.24], [0.01, 0.0, 0.28, 0.24], [0.01, 0.0, 0.28, 0.24], [0.01, 0.0, 0.28, 0.24], [0.01, 0.0, 0.28, 0.24], [0.01, 0.0, 0.28, 0.24], [0.01, 0.0, 0.28, 0.24], [0.01, 0.0, 0.28, 0.24], [0.01, 0.0, 0.28, 0.24], [0.01,'' \\
        \hline
    \end{tabular}
    \label{tab:example}
\end{table}

\begin{table}[h]
    \centering
    \caption{The best identified VQDv1 prompts for each model.}
    \label{tab:vqd-prompts}
    \begin{tabular}{@{}lp{9cm}@{}}
        \toprule
        \textbf{Model} & \textbf{Prompt} \\
        \midrule
        GPT-4o/GPT-4V & Generate a list of bounding box coordinates around the objects mentioned in the prompt if they exist in the image. Even if the prompt uses a singular verb like `is', generate multiple bounding boxes if multiple objects satisfy the query. The bounding box list should be formatted as: [[x\_min, y\_min, x\_max, y\_max]], and it can contain zero or more bounding boxes. Only provide the bounding box list, without any additional descriptions. \\
        LLaVA-NeXT & Please generate a list of bounding boxes coordinates of the region this query describes. Use the format [[x\_min,y\_min,x\_max,y\_max]....]. Do not respond in sentences, and only generate the bounding boxes. Respond with an empty list [[]], if no such region exists in the image. \\
        LLaVA (7B)/(13B) & Please answer the question by generating a list of bounding box coordinates around the objects the question is asking, and if no such object exists in the image, answer: [[]]\\
        \bottomrule
    \end{tabular}
\end{table}

\section{Model Details}
In this paper, all the open source MLLMs are loaded directly from HuggingFace, the detail models are below:
\begin{table}[ht]
\centering
\caption{MLLM Model Repository Paths}
\label{tab:model_paths}
\begin{tabular}{|>{\ttfamily}l|>{\ttfamily}l|}
\hline
\textbf{Model}     & \textbf{Repository Path}                             \\ \hline
LLaVA-NeXT         & llava-hf/llava-v1.6-mistral-7b-hf                    \\ \hline
InstructBlip       & Salesforce/instructblip-flan-t5-xxl                  \\ \hline
BLIP2              & Salesforce/blip2-flan-t5-xl                          \\ \hline
LLaVA1.5-7b        & llava-hf/llava-1.5-7b-hf                             \\ \hline
LLaVA1.5-13b       & llava-hf/llava-1.5-13b-hf                            \\ \hline
\end{tabular}
\end{table}

GPT-4v/4o are not open sourced, therefore we are unable to identify the models. We utilize the API released by OpenAI to evaluate four datasets on GPT-4v/4o.

\section{Additional Evaluation Details}

\textbf{Root Mean Squared Error (RMSE) Computation.} For TallyQA, besides Micro and Macro Accuracy, we also compute RMSE. However, we observed that due to the unpredictability of the MLLMs, the models occasionally output unreasonably large numbers as their predicted object counts. For instance, LLaVA-NeXT predicts an unreasonably large object count of 150 for one of the questions. Such outliers significantly inflate the models' overall average RMSE across all questions. As shown in the distribution of TallyQA questions, all counting numbers are between 0 and 15. Therefore, we apply a simple cutoff technique: an upper bound of 15 and a lower bound of 0 is applied to all predicted counts. This adjustment ensures that the RMSE remains meaningful and useful for analysis.

\textbf{Match Answer with Ground Truth}. \\
For TallyQA, the model is tasked with generating object counts. If the model correspondingly generates a number enclosed within a string, such as "2", we directly convert it to int type by type conversion. For the case where the model generates a word, we map the word to its corresponding number using the mappings shown in table ~\ref{tab:mapping}. Occasionally, the model generates answers that, while not numerical, still make sense. For example, the model might generate 'none' or 'no,' which we interpret as zero. We manually account for these cases and add additional mappings accordingly. While we acknowledge that even with these steps, we may still miss some unpredictable answers from the models, such as when the model responds with 'a few,' which is completely uninterpretable, we map these to None.

\begin{figure}[h]
\centering
\begin{tabular}{|c|c|c|c|c|c|}
\hline
\textbf{Word} & \textbf{Number} & \textbf{Word} & \textbf{Number} & \textbf{Word} & \textbf{Number} \\
\hline
zero & 0 & four & 4 & eight & 8 \\
none & 0 & five & 5 & nine & 9 \\
no & 0 & six & 6 & ten & 10 \\
one & 1 & seven & 7 & eleven & 11 \\
two & 2 & twelve & 12 & fourteen & 14 \\
three & 3 & thirteen & 13 & fifteen & 15 \\
\hline
\end{tabular}
\caption{Mapping of words to numbers in TallyQA}
\label{tab:mapping}
\end{figure}

In datasets like TallyQA, DVQA, and VQDv1, synonymous answers are rarely an issue due to the specific nature of each task. For example, TallyQA typically expects numerical answers that are definitive and unambiguous (numbers seldom have synonyms). The main exception is when 'none,' 'no,' and 'zero' are all interpreted as 0. In DVQA, which focuses on chart understanding, questions such as 'Which bar has the highest number?' require the model to read and provide the exact text from the graph, minimizing the possibility of synonymous answers. Similarly, VQDv1 involves generating bounding boxes and computing the Intersection over Union (IoU) to determine if the ground truth is correctly matched. The evaluation uses Recall and Precision metrics, which are not binary and therefore do not penalize synonymous answers.

In contrast, tasks in TDIUC are more likely to involve more interpretative answers. For example, the answers 'phone' and 'telephone' should be considered semantically similar and should both be acceptable if the ground truth is one of them. To minimize penalizing synonymous answers like the case above, we leverage WordNet\cite{WordNet}, a lexical database for the English language that is specifically designed for natural language processing. Specifically, we retrieve the sets of synonyms for each word from WordNet (using the synsets function) and compare these sets. If there is any overlap in the synsets, two words are considered synonyms, and we use this to evaluate if the predicted word(s) matches the ground truth(s).

\section{Additional Results}
\subsection{TallyQA}
\begin{figure}[H]
    \centering
    \includegraphics[width=\textwidth]{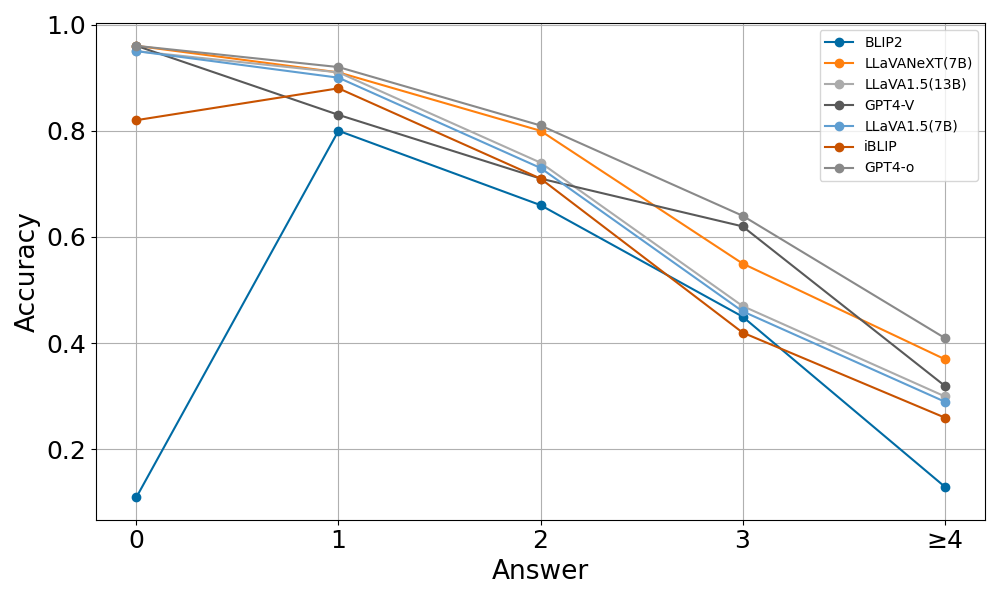}
    \caption{Accuracy as a function of the correct answer for simple counting questions in TallyQA.}
    \label{fig:tallyQA_simple_accuracy}
\end{figure}

\begin{figure}[H]
    \centering
    \includegraphics[width=\textwidth]{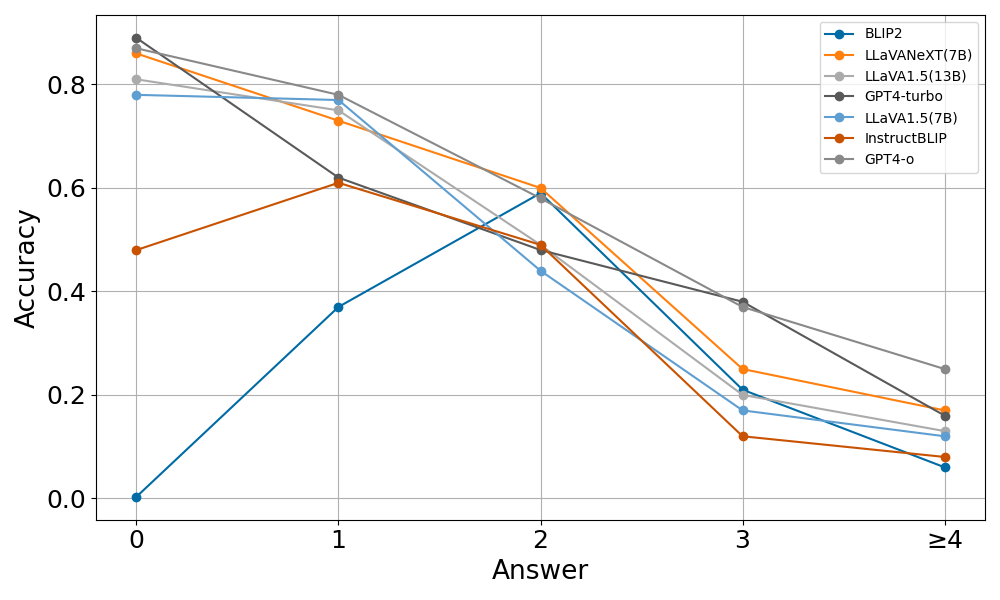}
    \caption{Accuracy as a function of the correct answer for complex counting questions in TallyQA.}
    \label{fig:tallyQA_complex_accuracy}
\end{figure}

For TallyQA, we found that the performance of most models decreases as the correct number to output increases, as shown in Figs.~\ref{fig:tallyQA_simple_accuracy} and \ref{fig:tallyQA_complex_accuracy}. Across counts, models perform much better at answering simple questions than complex questions.


\subsection{VQDv1}
\label{sec:vqd-appendix}

\paragraph{Alternative Prompts.}
All models performed poorly on VQDv1. As mentioned earlier, it was challenging to identify the best prompt for each model. We hypothesized that given the verbosity of GPT-4o, it would benefit from being allowed to provide more extended responses where it reasons `aloud.' However, this performed worse than the prompts used in our main results. In Table~\ref{tab:alternative-vqd-prompts}, we provide alternative prompts that we tried, where the results are given in Table~\ref{tab:alternative-vqd-results}.

\begin{table}[H]
    \centering
    \caption{Alternative prompts studied for VQDv1.}
    \label{tab:alternative-vqd-prompts}
    \begin{tabular}{@{}lp{9cm}@{}}
        \toprule
        \textbf{Model} & \textbf{Prompt} \\
        \midrule
        GPT-4o & Please generate a list of bounding boxes coordinates for regions that match what is described in the query. Bounding boxes should use the format: [[x\_min,y\_min,x\_max,y\_max], ..], where (x\_min,y\_min) is top left coordinate,(x\_max,y\_max) is bottom right coordinate. If there are no objects in the image that the query describes, please respond with an empty list. You can explain your answers if necessary, but end your response with the format: The bounding boxes coordinates are <box>[[x\_min,y\_min,x\_max,y\_max],..,..]<box>\". Please keep the special token <box> in your response.\\ \midrule
        LLaVA-NeXT & Generate a list of bounding box coordinates around the objects mentioned in the query, if they exist in the image. Even if the query uses a singular verb like `is', generate multiple bounding boxes if multiple objects satisfy the query. The bounding box list should be formatted as: [[x\_min, y\_min, x\_max, y\_max]], and it can contain zero or more bounding boxes. Only provide the bounding box list, without any additional descriptions. \\ \midrule
        LLaVA (7B)/(13B) & Generate a list of bounding box coordinates around the objects that the prompt mentioned if they exist in the image. Even if the query uses a singular verb like `is', you should still generate multiple bounding boxes if multiple objects satisfy the prompt. The bounding box list should be in the following format: [[x\_min, y\_min, x\_max, y\_max], [x\_min,y\_min, x\_max, y\_max]].\\
        \bottomrule
    \end{tabular}
\end{table}

\begin{table}[H]
\centering
\caption{MLLM performance on VQDv1 using the alternative prompts from Table~\ref{tab:alternative-vqd-prompts}.}
\label{tab:alternative-vqd-results}
\begin{tabular}{lccccc}
\toprule
Model & LLaVA (7B) & LLaVA (13B) & LLaVA-NeXT & GPT-4o\\
\midrule
Micro $F_1$ & 4.27\% & 8.90\% & 14.66\% & 23.81\%\\ 
\bottomrule
\end{tabular}
\end{table}

\paragraph{Qualitative Examples.} Among all the datasets we evaluated, all models consistently performed poorly on VQDv1. Consequently, we provide qualitative examples from VQDv1 in the figures below, using the prompts employed in our main results. These visualizations demonstrate the challenges models face when required to detect multiple objects.

\begin{figure}[h!]
    \centering
    \includegraphics[width=\textwidth]{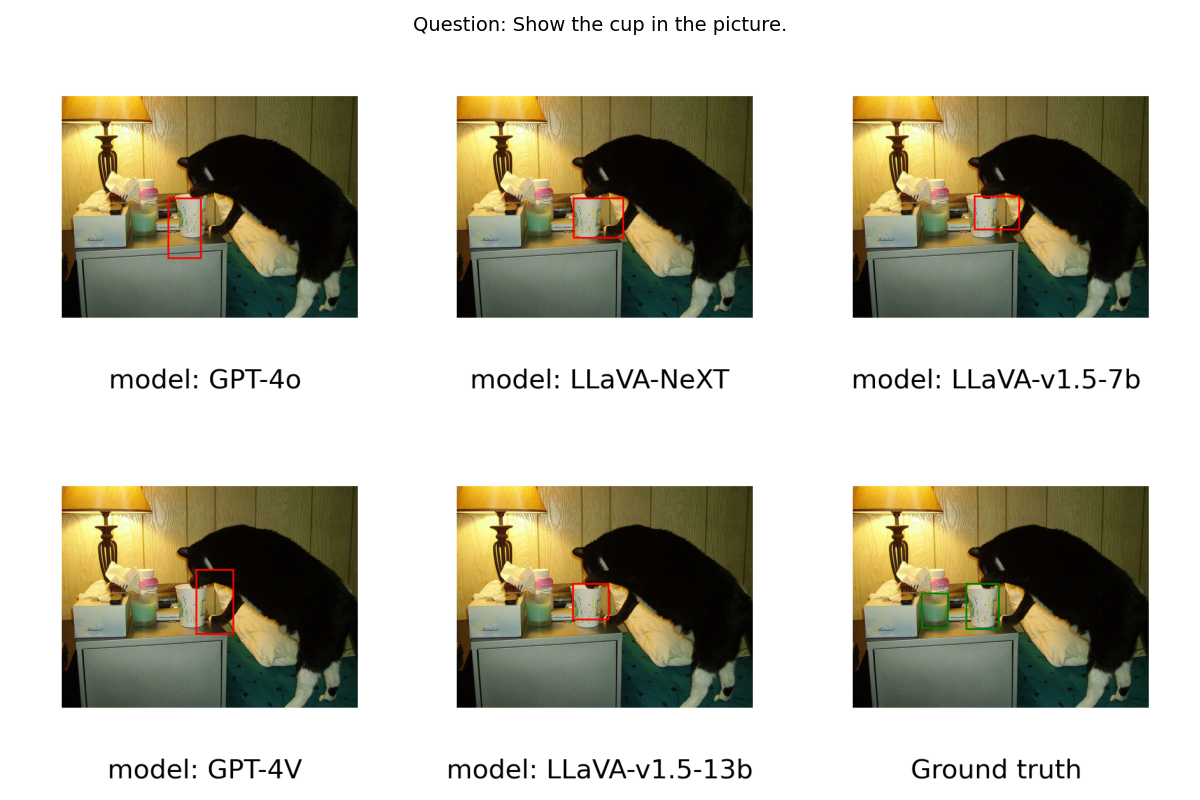}
\end{figure}

\begin{figure}[h!]
    \centering
    \includegraphics[width=\textwidth]{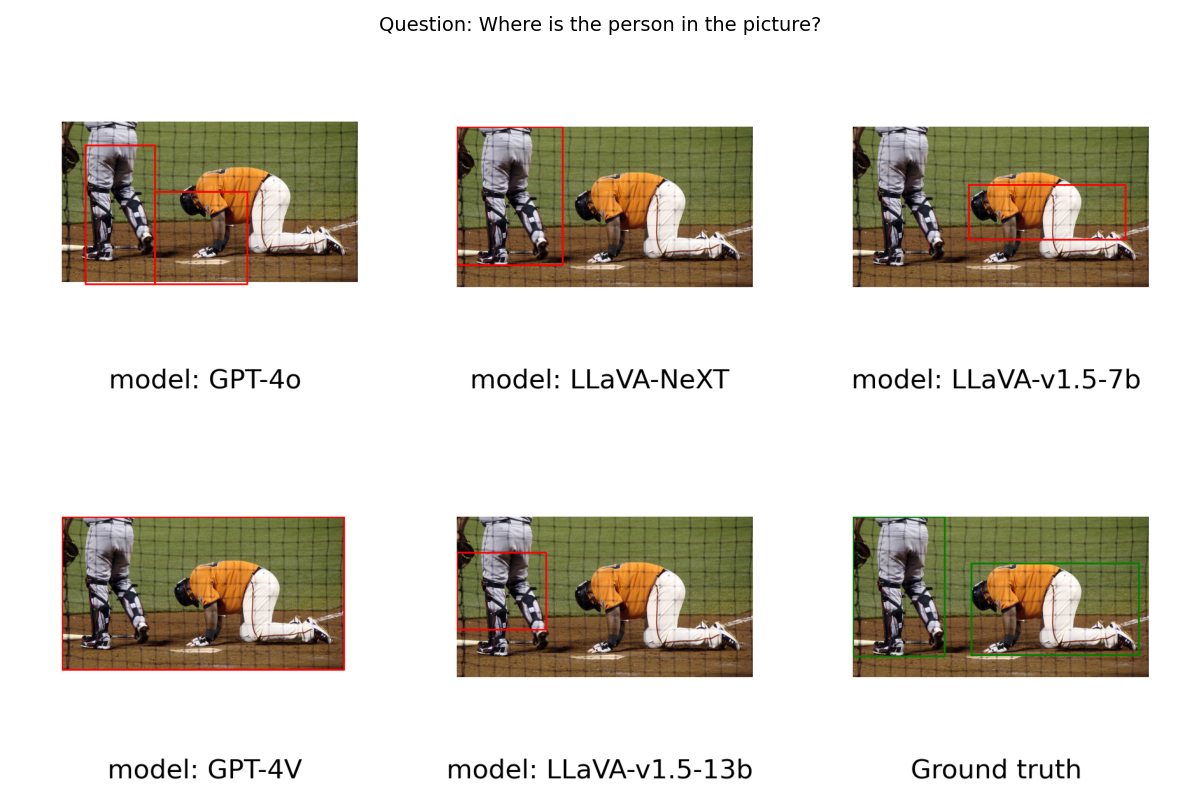}
\end{figure}
\begin{figure}[h!]
    \centering
    \includegraphics[width=\textwidth]{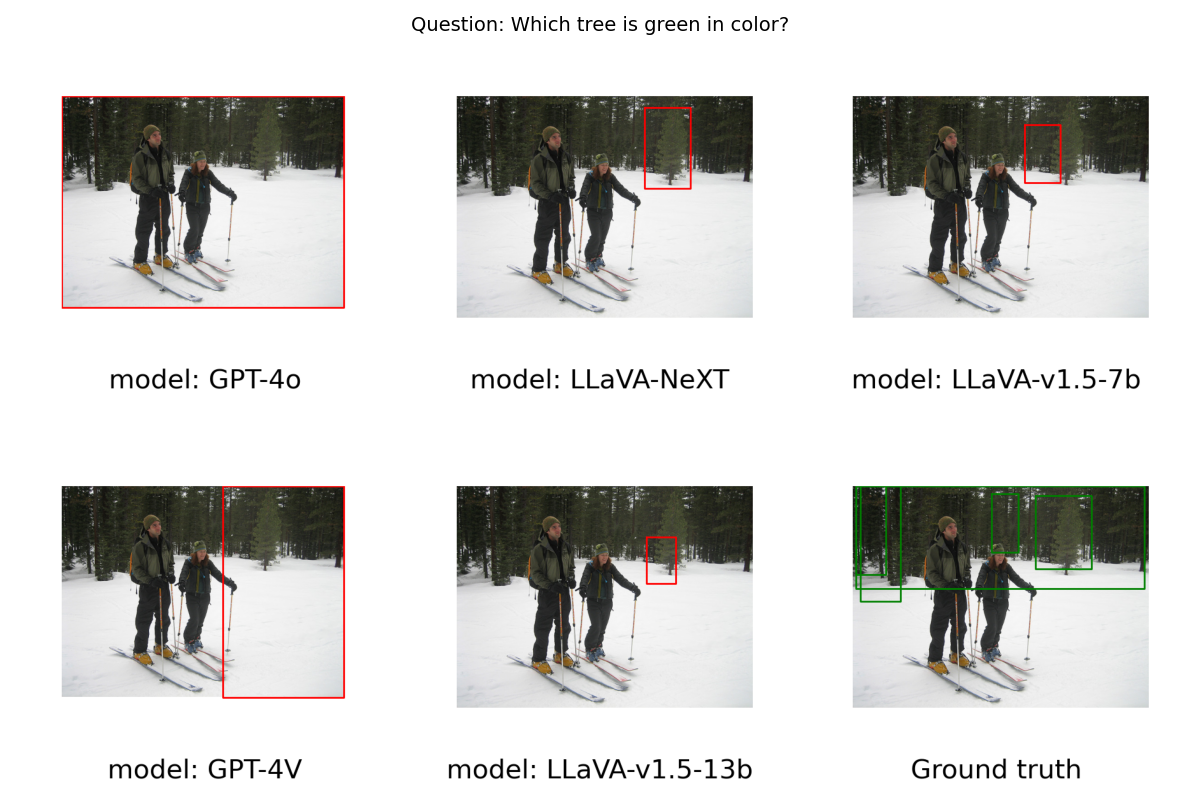}
\end{figure}

\end{document}